\documentclass[a4paper]{article}
\usepackage{epsfig}
\usepackage{subeqn}
\newcommand{\gbf}[1] {\mbox{\boldmath${#1}$\unboldmath}}
\newcommand{\be}{\begin{equation}}
\newcommand{\ee}{\end{equation}}
\newcommand{\negr}[1]{{\bf {#1}}}
\newcommand{\bseq}{\begin{subequations}}
\newcommand{\eseq}{\end{subequations}}
\newcommand{\beqa}{\begin{eqnarray}}
\newcommand{\eeqa}{\end{eqnarray}}
\usepackage{setspace}
\begin{document}
\date{}
\title{An Interval Analysis Based Study for the Design and the Comparison of 3-DOF Parallel Kinematic Machines}
\author{D. Chablat$^1$ Ph. Wenger$^1$ F. Majou$^1$ J-P. Merlet$^2$\\
 $^1$Institut de Recherche en Communications et Cybern\'etique de
Nantes
 \footnote{IRCCyN: UMR n$^\circ$ 6597 CNRS, Ecole Centrale de Nantes,
                     Universit\'e de Nantes, Ecole des Mines de Nantes} \\
 1, rue de la No\"e, 44321 Nantes, France \\
 $^2$INRIA Sophia-Antipolis \\
  2004 Route des Lucioles, 06902 Sophia Antipolis, France \\
  {\tt Damien.Chablat\symbol{64}irccyn.ec-nantes.fr} \\
  {\tt Philippe.Wenger\symbol{64}irccyn.ec-nantes.fr} \\
  {\tt F\'elix.Majou\symbol{64}irccyn.ec-nantes.fr} \\
  {\tt Jean-Pierre.Merlet\symbol{64}sophia.inria.fr}}
\maketitle
\section*{Abstract}
This paper addresses an interval analysis based study that is
applied to the design and the comparison of 3-DOF parallel
kinematic machines. Two design criteria are used, (i) a regular
workspace shape and, (ii) a kinetostatic performance index that
needs to be as homogeneous as possible throughout the workspace.
The interval analysis based method takes these two criteria into
account: on the basis of prescribed kinetostatic performances, the
workspace is analysed to find out the largest regular dextrous
workspace enclosed in the Cartesian workspace. An algorithm
describing this method is introduced. Two 3-DOF translational
parallel mechanisms designed for machining applications are
compared using this method. The first machine features three fixed
linear joints which are mounted orthogonally and the second one
features three linear joints which are mounted in parallel. In
both cases, the mobile platform moves in the Cartesian $x-y-z$
space with fixed orientation.
\par
{\bf Keywords:} Parallel kinematic machine, Design, Interval
analysis, Comparison, Workspace, Transmission factors.
\section{Introduction}
Parallel kinematic machines (PKM) are known for their high dynamic
performances and low positioning errors. The kinematic design of
PKM has drawn the interest of several researchers. The workspace
is usually considered as a relevant design criterion
\cite{Merlet:96,Clavel:88,Gosselin:91}. Parallel singularities
\cite{Wenger:97ICAR} occur in the workspace where the moving
platform cannot resist any effort. Thus are very undesirable and
generally eliminated by design. The Jacobian matrix, which relates
the joint rates to the output velocities is generally not constant
and not isotropic. Consequently, the performances (e.g. maximum
speeds, forces, accuracy and stiffness) vary considerably for
different points in the Cartesian workspace and for different
directions at one given point. This is a serious drawback for
machining applications \cite{Kim:1997,Treib:1998,Wenger:1999b}.
Few parallel mechanisms are isotropic throughout the workspace
\cite{Kong:2002,Carricato:2002}. But their low structural
stiffness make them inadequate for machining applications because
their legs are subject to bending.

To be of interest for machining applications, a PKM should
preserve good workspace properties, that is, regular workspace
shape and acceptable kinetostatic performances throughout. For
example in milling applications, the machining conditions must
remain constant along the whole tool path
\cite{Rehsteiner:1999a,Rehsteiner:1999b}. In many research papers,
this criterion is not taken into account in the algorithmic
methods used to compute the workspace volume
\cite{Luh:1996,Merlet:1999}. Other papers present methods that
compute the well-conditioned workspace using discretization
\cite{Stougthon:1993,Gosselin:1988}. Thus, the results they
provide cannot be proved formally. Conversely, interval analysis
methods applied to well-conditioned workspace computation provide
guaranteed results \cite{Merlet:2000,Chablat:2002ark}.

The comparison of PKM architectures is a difficult but relevant
challenge \cite{Wenger:1999b,joshi:2003}. Providing tools to allow
designers or end-users to rigorously compare PKM is indeed
necessary since the variety of existing PKM makes it hard to
choose which one is best-suited for a specific task.

In this paper, an interval analysis based method is addressed for
the design and comparison of 3-DOF PKM. This method takes into
account two criteria, (i) a regular workspace shape and, (ii) a
kinetostatic performance index that needs to be as homogeneous as
possible throughout the workspace. Two basic tools and an
algorithm that considers these two criteria are introduced: on the
basis of prescribed kinetostatic performances, the workspace is
analyzed to find out the largest regular dextrous workspace
(square, cube, cylinder, etc...) enclosed in the Cartesian
workspace.

Two translational parallel mechanisms derived from the Delta robot
\cite{Clavel:88} are compared using this method. The first
machine, called Orthoglide \cite{Chablat:2002}, features three
fixed linear joints which are mounted orthogonally and the second
one, called UraneSX (Renault Automation) \cite{Company:2000},
features three linear joints which are mounted in parallel. In
both cases, the mobile platform moves in the Cartesian $x-y-z$
space with fixed orientation.
 \par
Next section presents the interval analysis based method for 3-DOF
PKM design. Section~3 presents the Orthoglide and UraneSX
mechanisms, their kinematic equations and singularity analysis.
Section~4 reports the comparison between the two mechanisms
through the determination of the largest dextrous cube for the
Orthoglide and the largest dextrous square for the UraneSX
enclosed in the workspace.

\section{Description of the interval analysis based method for 3-DOF translational PKM design}

\subsection{Preliminaries}
\label{preliminaries}

\subsubsection{Dextrous Cartesian workspace}
For a 3-axis serial machine-tool, a parallelepiped-shaped
Cartesian workspace allows the end-user to visualize easily  where
to place cutting paths. This consideration should also hold for
PKM. However the workspace shape is often geometrically complex
and thus hard to visualize. Therefore, a regular-shaped workspace
is needed for PKM. Thus, we need to define a regular dextrous
workspace which is a regular-shaped workspace included in the
machine Cartesian workspace. Throughout the dextrous workspace, a
kinetostatic performance index (that is chosen beforehand) remains
as homogeneous as possible. This index can be the local or global
conditioning \cite{Gosselin:91} of the Jacobian matrix $\negr J$
(that maps the actuated joint rates of the manipulator into the
velocity of the mobile platform), the force or velocity
transmission factors. These last two indices make sense for 3-DOF
translational PKM with identical actuated joints.

The method presented in this paper aims at designing such 3-DOF
PKM. The velocity transmission factors are the ratio between the
actuated joints velocities and the velocity of the mobile
platform. They are the square roots $\psi_1$, $\psi_2$ and
$\psi_3$ of the real eigenvalues $\sigma_1$, $\sigma_2$ and
$\sigma_3$ of $(\negr J \negr J^T)^{-1}$. In order to keep
homogeneous kinetostatic properties, these factors are bounded
inside the dextrous workspace. The {\it regular dextrous Cartesian
workspace} can be defined as a set of points $P$ in the workspace
such that $\psi_1$, $\psi_2$ and $\psi_3$ are bounded, that is,
\begin{equation}
   {\cal W}_{Dextrous} = \{ P \in {\cal W} {~} | {~} \psi_{min} \leq
   \psi_i(P) \leq \psi_{max},~~i=1, 2, 3\}
   \label{equation:w_dextrous}
\end{equation}
Points $P$ in ${\cal W}_{Dextrous}$ are called {\it dextrous
points}.

The values of $\psi_{min}$ and $\psi_{max}$ (resp. $\sigma_{min}$
and $\sigma_{max}$) depend on given performance requirements. The
method described further aims at computing the largest dextrous
Cartesian workspace included in the Cartesian workspace, so that
its ratio to the Cartesian workspace is the best one. To be of
real interest for milling applications, a PKM must indeed include
a large regular dextrous workspace in its Cartesian workspace.

\subsubsection{Introduction to ALIAS library}
An algorithm for the definition of the largest dextrous workspace
included in the Cartesian workspace is described in the following
sections. This algorithm uses the {\tt ALIAS}
library~\cite{Merlet:2000}, which is a C++ library of algorithms
based on interval analysis. These algorithms deal with systems of
equations and inequalities whose expressions are an arbitrary
combination of the most classical mathematical functions
(algebraic terms, sine, cosine, log etc..) and whose coefficients
are real numbers or, in some cases, intervals. An interface exists
with Maple that allows the automatic generation of C++ codes being
given the Maple description of the system and then to compile and
run the generated code in order to get the result within the Maple
session. Without being exhaustive,  {\tt ALIAS} library provides
algorithms that enable one to, (i) find an approximation of the
real roots of n-dimensional systems, (ii) find an approximation of
the variety defined by n-dimensional systems, (iii) find an
approximation of the global minimum or maximum of a function
(eventually under equations and/or inequalities constraints) up to
an accuracy provided by the user, (iv) analyze a system of
algebraic equations to determine bounds for its real roots.

\subsubsection{Geometric constraints}
The dextrous workspace ${\cal W}_{Dextrous}$ is defined in
Eq.~\ref{equation:w_dextrous} as a function of the eigenvalues of
 $(\negr J \negr J^T)^{-1}$. These eigenvalues are
determined by solving the third degree characteristic polynomial
${\cal P}$ of $(\negr J \negr J^T)^{-1}$. To decrease the
computing time and to avoid numerical problems on singularities,
it is recommended to add geometrical constraints. These
constraints naturally depend on the mechanism architecture (see \S
\ref{geometric_constraints}).

\subsection{A first basic tool: Box verification}
Our purpose is to determine the largest regular dextrous workspace
that is enclosed in the Cartesian workspace. For a given point, we
note {\it valid point} if it is a dextrous point and {\it invalid
point} otherwise. For that purpose we need to design first a
procedure, called ${\cal M}(B)$, that takes as input a Cartesian
box $B$ and returns:
\begin{itemize}
 \item 1: if every point in $B$ is valid,
 \item -1: if no point in $B$ is valid,
 \item 0: if neither of the other two conditions could be verified.
\end{itemize}
The first step of this procedure consists in considering an
arbitrary point of the box (e.g. its center) and to compute the
eigenvalues at this point: either all of them lie in the range
$[\sigma_{min},\sigma_{max}]$ in which case the center is called
{\em valid}  or at least one of them lie outside this range and
the center point is denoted {\em invalid}.
\subsubsection{Valid center point}
In that case if we are able to check that there is no point in $B$
such that one of the eigenvalues at this point is  equal to
$\sigma_{min}$ or $\sigma_{max}$, then we can guarantee that every
point in $B$ is valid.
Indeed assume that at a given point
$B$ the lowest eigenvalue is lower than $\sigma_{min}$: this
implies that somewhere along the line joining this point to the
center of the box the lowest eigenvalue is exactly $\sigma_{min}$.
 \par
To perform this check we set the unknown in the characteristic
polynomial ${\cal P}$ of $(\negr J \negr J^T)^{-1}$ to
$\sigma_{min}$ (and then to $\sigma_{max}$) and we get a
polynomial in $x, y, z$ only. We now have to determine if there
exists some values for these three Cartesian coordinates that
cancel the polynomial, being understood that these values have to
define a point belonging to $B$. This is done by using an interval
analysis algorithm from the {\tt ALIAS}
library~\cite{Merlet:2000}. The principle of this algorithm is to
calculate first the polynomial value for the center point $C_B$ of
$B$. Without lack of generality we may assume that this value is
positive. If we are able to determine a point $S_B$ in $B$ such
that the polynomial value at this point is negative, then we can
guarantee that there exists a point on the line joining $C_B$ to
$S_B$ such that the polynomial is exactly 0. The purpose of the
algorithm is now to determine if such a point exist. Now let $B_i$
be a box included in $B$: using interval analysis we are able to
calculate a range $[m_{B_i}, M_{B_i}]$ such that for any point $X$
in $B_i$ we have $m_{B_i} \le {\cal P}(X) \le M_{B_i}$. Note that
this interval evaluation is numerically safe as the bounds of the
range are calculated by taking into account round-off errors. On
the other hand these bounds may not be {\em sharp} i.e. there may
be no $X$ in $B_i$ such that ${\cal P}(X)=m_{B_i}$ or $M_{B_i}$.
Note, however, that  the width of the overestimation decreases
with the width of $B_i$. Furthermore we may get a sharp evaluation
by using, for instance, the derivatives of ${\cal P}$. Indeed we
may calculate the interval evaluation $[r_{x,y,z},R_{x,y,z}]$ of
$\partial {\cal P}/
\partial x,y,z$ and if all three interval evaluations have constant
signs (i.e. $r_{x,y,z}>0$ or $R_{x,y,z}<0$), then sharp $m_{B_i},
M_{B_i}$ are obtained by setting the variables to fixed values.
For instance if $r_{x}>0$, then $m_{B_i}(M_{B_i})$ is obtained by
setting $x$ to its lower (upper) bound. Note that other methods
may also be used to determine sharp bounds
(see~\cite{moore79-i,neumaier90-i,ratscheck95-i}).
 \par
Hence we have the following properties:
\begin{enumerate}
 \item if $m_{B_i}>0$, then for any point in $B_i$ the polynomial
 $\cal P$ is positive
 \item if $M_{B_i}<0$, then for any point in $B_i$ the
polynomial $\cal P$ is negative
 \item if $m_{B_i}<0$ and $M_{B_i}>0$
and the bounds are sharp, then the polynomial $\cal P$ cancels in
$B_i$
 \item if $m_{B_i}<0$ and $M_{B_i}>0$ and the bounds are not sharp,
then we cannot guarantee the sign of the polynomial $\cal P$
within $B_i$
\end{enumerate}
At that point a simple branch-and-bound algorithm is used: the
initial box $B$ is bisected until either all the sub-boxes
resulting from the bisection satisfy property 1 (in which case we
can guarantee that the polynomial $\cal P$ never cancels for $B$
and consequently that all the eigenvalues of ${\cal P}$ lie in the
range $[\sigma_{min},\sigma_{max}]$, which implies ${\cal
M}(B)=1$) or a sub-box resulting from the bisection satisfies
property 2 or 3 which means that at some point in $B$ at least one
of the eigenvalues of ${\cal P}$ lies outside the range
$[\sigma_{min}, \sigma_{max}]$, which corresponds to ${\cal
M}(B)=0$.

The algorithm may indeed return $0$ for a box that includes only
valid points. But the width of this box will be lower than
$\alpha/2$ (where $\alpha$ is an accuracy threshold fixed in
advance for the computation)
and hence the final result will be within the tolerance margin of
the calculation. The only case in which the calculation will be
not guaranteed will occur only when $\alpha$ is lower than the
machine accuracy. But we may determine that we are in such
configuration as the width of the box from the machine viewpoint
will be 0: if a box of width 0 is processed and the algorithm
returns 0, then a warning message will be issued indicating that
the calculation is no more guaranteed. Note, however, that we may
still use the algorithm by using a multi-precision package such as
MPFR that will allow to get a guaranteed result. Furthermore it is
doubtful that computing the result with an accuracy better than
the machine precision makes sense.
\subsubsection{Invalid center point}
Without lack of generality, we may assume that at the center of
the box the largest eigenvalue is greater than $\sigma_{max}$. If
there is no point in $B$ such that one of the eigenvalues is equal
to $\sigma_{max}$, then we can guarantee that for any point in $B$
the largest eigenvalue is always greater than $\sigma_{max}$ and
consequently ${\cal M}(B)= -1$. This check is performed by using
the same method as in the previous case.
\subsection{A second basic tool: Box workspace verification}
During the calculation of the dextrous workspace, we consider a
Cartesian box $B$ and we have to examine if this box may contain a
point that is the center of a Cartesian box $B_W$ with edge length
$w$, which is fully enclosed in the robot workspace. We assume
here that this workspace is defined by a set of $m$ inequalities
$F_j$ such that a point $X$ belongs to the workspace if $F_j(X)\le
0$ for all $j$ in $[1,m]$. Let $B_i$ be a sub-box included in $B$,
defined by the three ranges $[\underline{x_i},\overline{x_i}]$,
$[\underline{y_i},\overline{y_i}]$,
$[\underline{z_i},\overline{z_i}]$. All the boxes with edge length
$w$ that have as center a point in $B_i$ are included in the {\em
hull box} $H_{B_i}$ defined by the three ranges
$[\underline{x_i}-w/2,\overline{x_i}+w/2]$,
$[\underline{y_i}-w/2,\overline{y_i}+w/2]$,
$[\underline{z_i}-w/2,\overline{z_i}+w/2]$. As in the previous
section we may use interval analysis to compute an interval
evaluation $[m^j_{B_i}, M^j_{B_i}]$ of all $F_j(H_{B_i})$ with the
following properties:
\begin{enumerate}
\item if $M^j_{B_i}<0$ for all $j$ in $[1,m]$, then any point of $B_i$
may be the center of a box with edge length $w$ that is included in
the workspace
\item if $m^j_{B_i}>0$, then no  point of $B_i$
may be the center of a box with edge length $w$ that is included in
the workspace
\item if $m_{B_i}<0$ and $M_{B_i}>0$, then we cannot determine if some
point within $B_i$ may be the center of $B_W$
\end{enumerate}
Note also that if  the widths of all the ranges defining $B_i$ are
lower than $w$, any box $B_W$ contains the 4 corners of the box
$B_i$.
 \par
Using a similar branch-and-bound algorithm as in the previous
section, we may now determine if either all, none or some points
of $B$ may be the center of a box $B_W$. The initial box $B$ is
bisected until either all the sub-boxes resulting from the
bisection satisfy property 1 (then any point of $B$ may be the
center of a box $B_W$), or 2 (no point of $B$ may be the center of
a box $B_W$). If a sub-box satisfies property 3 and the widths of
its ranges are lower than $w$, we check if the corners of $B$
belong to the workspace: if all the corners either belong or do
not belong to the workspace we continue the bisection. If we have
a mixed situation with some corners belonging to the workspace
whereas other ones do not, we may state that $B$ contains both
points that may be the center of a box $B_W$ and points that
cannot. A similar situation is obtained if we have found at least
a sub-box that satisfies property 1 and a sub-box that satisfies
property 2.
 \par
At that point we may define a procedure ${\cal G}(B,w)$ that takes as input
a box $B$ and an edge length $w$ and returns:
\begin{itemize}
 \item -1: there is no points in $B$ that may be the center of a box
$B_W$
 \item 1: all the points in $B$ may be the center of a box $B_W$
 \item 0: $B$ contains both points that may be the center of a box
$B_W$ and points that cannot.
\end{itemize}

\subsection{Algorithm for the determination of a cubic dextrous Cartesian workspace}
\label{algorithm}

An algorithm is now described for the determination of a cube that
is enclosed in the Cartesian workspace and aligned with the
coordinate axis, whose edge length is $2w$ and such that there is
no other cube enclosed in the workspace with an edge length of
$2(w+\alpha)$. 
This algorithm can be applied to any 3-DOF manipulator.
Other shapes for regular dextrous workspace is considered in
section \ref{other_shapes}.
 \par
The first step is to determine the largest cube enclosed in the
workspace with a center located at $(0, 0, 0)$. This is done by
using the ${\cal M}$ procedure on the Cartesian box $B_{init}$
$[-k\alpha, k\alpha],[-k\alpha, k\alpha], [-k\alpha, k\alpha]$
where $k$ is an integer initialized to 1. Each time the ${\cal M}$
procedure returns 1 for $B_{init}$ (which means that the cube with
edge length $2k\alpha$ is enclosed in the dextrous workspace) we
double the value of $k$. If this procedure returns -1 for a value
of $k$ larger than 1 this implies that the cube with edge length
$k\alpha/2$ is in the dextrous workspace whereas the cube with
edge length $k\alpha$ is not. Hence if $k>2$ (otherwise no
improvement is possible) we restart the process with
$k=(k/2+k)/2$. After a failure at $k_{fail}$ the principle is to
always choose a value of $k$ which is the mid-point between the
last value $k_s$ of $k$ for which ${\cal M}=1$ and $k_{fail}$
until $k_{fail}=k_s+1$.   For example if ${\cal M}$ returns 1 for
$k$=1, 2, 4 and returns -1 for $k=8$ we set $k$ to 6. Otherwise we
have determined that the cube with edge length $2k\alpha$ is
enclosed in the dextrous workspace, whereas the cube with edge
length $2(k+1)\alpha$ is not. The value $2k\alpha$ is hence an
initial value for $w$. Note that the above procedure may be used
whatever the coordinates of the center: it is implemented as a
general purpose procedure ${\cal C}(x_M, y_M, z_M)$ that takes as
input the coordinates of a point $M$ and returns the edge length
of the largest cube centered at $M$, that is enclosed in the
dextrous workspace.
\par
In the algorithm for determining the largest cube enclosed in the
dextrous workspace, we manage a list of Cartesian boxes ${\cal L}$
that are processed by the algorithm in sequence. During the
processing, boxes may be added to a list. At one step of the
algorithm we have $n$ boxes in the list whereas processing box
numbered $i$ (which means that boxes numbered from 0 to $i-1$ have
already been processed and may be discarded whereas boxes $i$ to
$n$ have to be processed). The algorithm stops when all the boxes
in ${\cal L}$ have been processed. The box numbered $i$ in the
list is denoted $B_i$ and the maximum number of boxes in ${\cal
L}$ is $N$.
\par
At the beginning of the algorithm, ${\cal L}$ has only one box
$B_0$ that contains the workspace (for example for the Orthoglide
$B_0$= \{$[-L, L]$, $[-L, L]$, $[-L, L]$\}). The algorithm can be
described by the following six steps:
\begin{enumerate}
 \item calculate $w=C(0,0,0)$ 
 \item if $i > n$ EXIT 
 \item if ${\cal G}(B_i,w+alpha)$= -1, then set $i$ to $i+1$ and go to 2 
 \item if ${\cal G}(B_i,w+alpha)$= 1, then calculate $w^\prime=C(x_{B_i},
y_{B_i},z_{B_i})$ where $x_{B_i},
y_{B_i},z_{B_i}$ are the coordinates of the center of $B_i$. If
$w^\prime>w$, then update $w$. Go to step 6 
\item if ${\cal G}(B_i,w+alpha)$= 0, then go to step 6 \item
bisect the variable in the box $B_i$ that has the largest range.
For example if the box $B_i$ is defined as
$[\underline{x_i},\overline{x_i}]$,
$[\underline{y_i},\overline{y_i}]$,
$[\underline{z_i},\overline{z_i}]$ and the variable $x$ has the
largest range the bisection process creates two new boxes $B^1_i
=\{[\underline{x_i},(\underline{x_i}+\overline{x_i})/2],
[\underline{y_i},\overline{y_i}],
[\underline{z_i},\overline{z_i}]\}$ and $B^2_i
=\{[(\underline{x_i}+\overline{x_i})/2,\overline{x_i}],
[\underline{y_i},\overline{y_i}],
[\underline{z_i},\overline{z_i}]\}$. If $n<N/2$, both boxes are
stored at the end of the list (and we set $i$=$i$+1), otherwise
 Box $B^1_i$ are stored in
${\cal L}$ in place of $B_i$ whereas box $B^2_i$ is stored at
location $i+1$ after a shift of the boxes $B_{i+1},\ldots,B_n$.
Set $n$ to $n+1$ and go to 2.
\end{enumerate}

Step 1 allows one to establish an initial value for the maximal
edge length. Step 3 eliminates boxes that cannot contain the
center of the maximal cube due to the workspace limits. Boxes
satisfying step 4 are candidate to include the center of the
largest cube: hence we calculate the largest cube centered at the
box that may allow to update the current value of the largest
edge. Step 6 is the bisection process that allows one to decrease
the size of the box with the effect of a sharper calculation for
the procedure ${\cal G}$. Note also that two storage modes that
are used for adding the boxes resulting from the bisection
process. The second mode allows for a minimal memory storage but
has the drawback of focusing on a given part of the workspace
whereas the center of the largest cube may be located in another
part. The first mode makes it possible to explore various parts of
the workspace which may result in large improvement on the value
of $w$ but as the drawback of possibly creating a large number of
boxes. The proposed storage mode allows one to mix the advantages
of both storage modes.

This procedure ensures to determine a cube with edge length $w$
that is enclosed in the workspace and in the dextrous workspace,
whereas there is no such cube with edge length $w+\alpha$.

Note that an incremental approach is possible. After having
computed $w=w_1$ with a given accuracy $\alpha$ it is always
possible to replace the initial value of $w$ as calculated in step
1 of the algorithm by the value $w_1$ when computing the cube with
a lower value for $\alpha$. Computation times of the largest cube
for various accuracies are given for a specific 3-DOF PKM in
section \ref{results}.

\subsection{Other regular dextrous workspace shapes}
\label{other_shapes}

Clearly, considering the largest cube may not be appropriate if
the studied PKM has a rectangular or a spherical-shaped workspace.
The algorithm can thus be modified. Here are for example the
necessary changes that must be taken into account to consider the
largest sphere: the idea is to use spherical coordinates and hence
$x, y, z$ are substituted by $x_c+r\sin\psi\sin\theta,
y_c+\cos\psi\sin\theta, z_c+r\cos\theta$, with $r$ in [0,R],
$\psi, \theta$ in $[0,2\pi]$, $x_c, y_c, z_c$ being the
coordinates of the center of the sphere and R its radius. Interval
analysis allows to deal with expressions involving sine and cosine
and hence procedures ${\cal M}, {\cal G}$ can still be used with
these new parameters. Similarly procedure ${\cal C}(x_M, y_M,
z_M)$ can be used to determine the largest radius of the sphere
centered at $(x_M, y_M, z_M)$ for which the eigenvalues are valid.
Hence, with this modification, the algorithm can calculate the
largest sphere enclosed in the dextrous workspace.

Spheres and cubes are defined by their center and one additional
parameter. But other shapes may involve more parameters: for
example a cylinder needs a center but also a height and a radius.
We can still perform a change in the variables so that procedures
${\cal M}, {\cal G}$ can still be used. The key point is that
procedure ${\cal C}(x_M, y_M, z_M)$ has to be modified as we have
now two optimization parameters. But in that case volume
optimization alone has less meaning: for example the optimization
result for a cylinder may be a cylinder with a relatively small
radius and a large height, which may be of no interest. A cylinder
of identical radius and height with a lower volume may be the most
interesting result. A possible way to manage this problem is to
assign a range $[a,b]$ for the ratio $R/h$ where $R$ is the
cylinder radius and $h$ its height. In that case the procedure
${\cal C}$ has to solve an optimization problem which is to
maximize the volume of the cylinder under the constraints that the
eigenvalues are valid and the ratio $R/h$ satisfies $a \le R/h \le
b$. {\tt ALIAS} is still able to manage such an optimization
procedure.

\subsection{Approximate calculation of the dextrous workspace}
Small modifications in the previous algorithm allow to determine
an approximation of the dextrous Cartesian workspace ${\cal
W}_{Dextrous}$ as a set ${\cal S}$ of 3D Cartesian boxes such that
for any box $B$ in ${\cal S}$ and for any point in $B$ the
constraints on the eigenvalues are satisfied. The width of all the
boxes in the set ${\cal S}$ is greater than a given threshold
$\epsilon$: hence we get only an approximation of the dextrous
Cartesian workspace. But the algorithm provides the volume $V_a$
of the approximation and a volume error $V_e$ such that the volume
$V_d$ of the  dextrous Cartesian workspace satisfies $V_d \le
V_a+V_e$. Decreasing the value of $\epsilon$ makes it possible to
increase $V_a$ and to decrease $V_e$. In this paper, this method
is used to analyze 3D boxes but it can be applied for any
mechanism with $n$ d.o.f., the result being a set of $n$D boxes.

Initially $V_a, V_e$ are set to 0.
\begin{enumerate}
\item if $i > n$ EXIT 
\item if ${\cal M}(B_i)$= -1, then set $i$ to $i+1$ and go to 1 
\item if ${\cal M}(B_i)$= 1, then store $B_i$ in ${\cal S}$ and
add
its volume to $V_a$. Set $i$ to $i+1$ and go to 1 
\item if ${\cal M}(B_i)$= 0, then
    \begin{enumerate}
    \item if the largest width of $B_i$ is lower than $\epsilon$,
then add its volume to $V_e$, set $i$ to $i+1$ and go to 1
    \item otherwise go to step 5
    \end{enumerate} 
\item process bisection for the box $B_i$. Set $n$ to $n+1$ and go
to 1
\end{enumerate}
Note that this procedure may be incremental if the boxes neglected
at step 4-a are stored in a file ${\cal F}$. Indeed a first run
with a given $\epsilon$ allows to obtain initial values for $V_a,
V_e$. If the quality of the approximation is not satisfactory, we
may choose a smaller value of $\epsilon$ (e.g. $\epsilon$/2). But
instead of starting with the initial $B_0$, we may use the boxes
stored in ${\cal F}$, thereby avoiding to repeat computation that
has already been done during the initial run.

\section{Description of the Orthoglide and the UraneSX}

The previous interval analysis based design method is now applied
to the comparison of two 3-DOF translational PKM. It is
particularly interesting to compare these two mechanisms because
they belong to the same architecture family.

\subsection{Orthoglide and UraneSX architectures}
Most existing PKM can be classified into two main families. PKM of
the first family have fixed foot points and variable length struts
and are generally called ``hexapods'' or ``tripods''. PKM of the
second family have variable foot points and fixed length struts.
They are interesting because the actuators are fixed and thus the
moving masses are lower than in the hexapods and tripods.

\begin{figure}[!hb]
    \begin{center}
           \centerline{\hbox{\includegraphics[width=80mm,height=61mm]{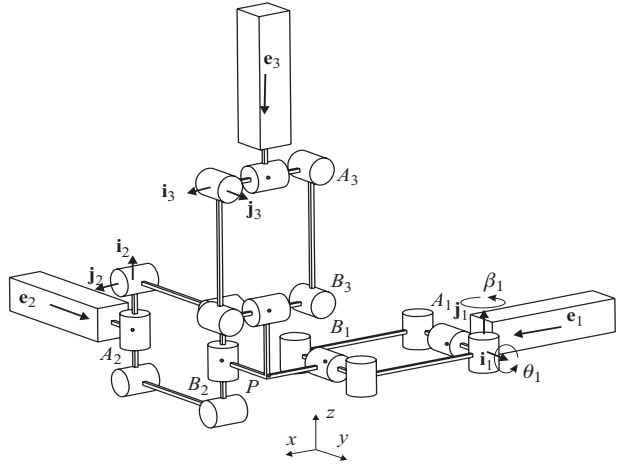}}}
           \caption{{\bf Orthoglide kinematic architecture}}
           \protect\label{figure:Orthoglide}
    \end{center}
\end{figure}

\begin{figure}[!ht]
    \begin{center}
           \centerline{\hbox{\includegraphics[width=46mm,height=57mm]{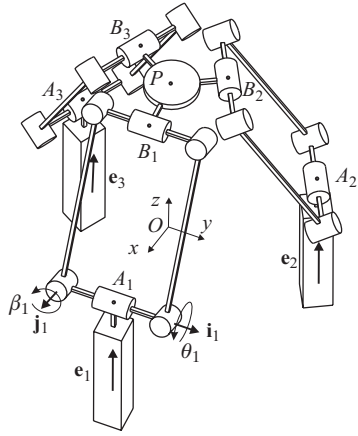}}}
           \caption{{\bf UraneSX kinematic architecture}}
           \protect\label{figure:UraneSX}
    \end{center}
\end{figure}

The Orthoglide and the UraneSX mechanisms studied in this paper
are $3$-DOF translational PKM and belong to the second family.
Figures~\ref{figure:Orthoglide} and \ref{figure:UraneSX} show the
general kinematic architecture of the Orthoglide and of the
UraneSX, respectively. Both mechanisms have three parallel $PRPaR$
identical chains (where $P$, $R$ and $Pa$ stand for Prismatic,
Revolute and Parallelogram joint, respectively). The actuated
joints are the three linear joints. These joints can be actuated
by means of linear motors or by conventional rotary motors with
ball screws.

The output body is connected to the linear joints through a set of
three parallelograms of equal lengths $L~=~A_iB_i$, so that it can
move only in translation. Vectors $\negr e_i$ coincide with the
direction of the $i$th linear joint. The base points $A_i$ are
located at the middle of the first two revolute joints of the
$i^{th}$ parallelogram, and $B_i$ is at the middle of the last two
revolute joints of the $i^{th}$ parallelogram.

For the Orthoglide mechanism, the first linear joint axis is
parallel to the $x$-axis, the second one is parallel to the
$y$-axis and the third one is parallel to the $z$-axis. When each
vector $\negr e_i$ is aligned with $\negr A_i \negr B_i$, the
Orthoglide is in an isotropic configuration and the tool center
point $P$ is located at the intersection of the three linear joint
axes.

The linear joint axes of the UraneSX mechanism are parallel to the
$z$-axis. In fig.~\ref{figure:UraneSX}, points $A_1$, $A_2$ and
$A_3$ are the vertices of an equilateral triangle whose geometric
center is $O$ and such that $OA_i=R$. Thus, points $B_1$, $B_2$
and $B_3$ are the vertices of an equilateral triangle whose
geometric center is $P$, and such that $OB_i=r$.

\subsection{Kinematic equations and singularity analysis}
We recall briefly here the kinematic equations and the
singularities of the Orthoglide and of the UraneSX (See
\cite{Company:2000,Chablat:2002} for more details).

Let $\theta_i$ and $\beta_i$ denote the joint angles of the
parallelogram about  axes $\negr i_i$ and $\negr j_i$,
respectively (Figs.~\ref{figure:Orthoglide} and
\ref{figure:UraneSX}). Let $\negr \rho_1$, $\negr \rho_2$, $\negr
\rho_3$ denote the linear joint variables and $L$ denote the
length of the three legs, $A_iB_i$.

For the Orthoglide, the position vector \negr p of the tool center
point $P$ is defined in a reference frame (O, $x$, $y$, $z$)
centered at the intersection of the three linear joint axes (note
that the reference frame has been translated in
Fig.~\ref{figure:Orthoglide} for more legibility).

For the UraneSX, the position vector \negr p of the tool center
point $P$ is defined in a reference frame (O, $x$, $y$, $z$)
centered at the geometric center of the points $A_1$, $A_2$, and
$A_3$ (same remark as above).

Let $\dot{\gbf {\rho}}$ be referred to as the vector of actuated
joint rates and $\dot{\negr p}$ as the velocity vector of point
$P$:
 \be
    \dot{\gbf{\rho}}=
    [\dot{\rho}_1~\dot{\rho}_2~\dot{\rho}_3]^T
   ,\quad
    \dot{\negr p}=
    [\dot{x}~\dot{y}~\dot{z}]^T
 \ee
$\dot{\negr p}$ can be written in three different ways by
traversing the three chains $A_iB_iP$:
\begin{equation}
    \dot{\negr p} =
    \negr e_i \dot{\rho}_i +
    (\dot{\theta}_i \negr i_i + \dot{\beta}_i \negr j_i)
    \times
    (\negr b_i - \negr a_i) \\
   \label{equation:cinematique}
\end{equation}
where $\negr a_i$ and $\negr b_i$ are the position vectors of the
points $A_i$ and $B_i$, respectively, and $\negr e_i$ is the
direction vector of the linear joints, for $i=1, 2, $3.

We want to eliminate the three idle joint rates $\dot{\theta}_i$
and $\dot{\beta}_i$ from Eqs.~(\ref{equation:cinematique}), which
we do by dot-product of Eqs.~(\ref{equation:cinematique}) by
$\negr b_i - \negr a_i$:
\begin{equation}
   (\negr b_i - \negr a_i)^T \dot{\negr p} =
   (\negr b_i - \negr a_i)^T
   \negr e_i \dot{\rho}_i
 \label{equation:cinematique-2}
\end{equation}
Equations (\ref{equation:cinematique-2}) can now be cast in vector
form, namely $\negr A \dot{\bf p} = \negr B \dot{\gbf \rho}$,
where \negr A and \negr B are the parallel and serial Jacobian
matrices, respectively:
\begin{eqnarray}
   \negr A =
   \left[\begin{array}{c}
           (\negr b_1 - \negr a_1)^T \\
           (\negr b_2 - \negr a_2)^T \\
           (\negr b_3 - \negr a_3)^T
         \end{array}
   \right]
   {\rm ~~and~~}
   \negr B =
   \left[\begin{array}{ccc}
            \eta_1&
            0 &
            0 \\
            0 &
            \eta_2&
            0 \\
            0 &
            0 &
            \eta_3
         \end{array}
   \right]
 \label{equation:A_et_B}
\end{eqnarray}
with $\eta_i= (\negr b_i - \negr a_i)^T \negr e_i $ for $i=1,2,3$.
\par
Parallel singularities occur when the determinant of the matrix
\negr A vanishes, {\it i.e.} when $\det(\negr A)=0$.
Eq.~(\ref{equation:A_et_B}) shows that the parallel singularities
occur when:
 \be
    (\negr b_1 - \negr a_1) =
    \alpha  (\negr b_2 - \negr a_2) +
    \lambda (\negr b_3 - \negr a_3)
 \ee
that is when the points $A_1$, $B_1$, $A_2$, $B_2$, $A_3$ and
$B_3$ lie in parallel planes. A particular case occurs when the
links $A_iB_i$ are parallel:
\begin{eqnarray}
    (\negr b_1 - \negr a_1) &||&
    (\negr b_2 - \negr a_2)
    \quad {\rm and} \nonumber \\
    (\negr b_2 - \negr a_2) &||&
    (\negr b_3 - \negr a_3)
    \quad {\rm and} \nonumber \\
    (\negr b_3 - \negr a_3) &||&
    (\negr b_1 - \negr a_1)\nonumber
\end{eqnarray}
Serial singularities arise when the serial Jacobian matrix \negr B
is no longer invertible {\it i.e.} when $\det(\negr B)=0$. At a
serial singularity a direction exists along which no Cartesian
velocity can be produced. Equation~(\ref{equation:A_et_B}) shows
that $\det(\negr B)=0$ when for one leg $i$, $(\negr b_i - \negr
a_i) \perp \negr e_i$. 

When \negr B is not singular, we can write,
\begin{equation}
   \dot{\gbf \rho} = \negr J^{-1}  \dot{\negr p}
   {\rm ~with~}
   \negr J^{-1} = \negr B^{-1} \negr A
\end{equation}

\section{Comparison of the Orthoglide and the UraneSX}
In this section, we calculate the edge length of the largest cube
for the Orthoglide, the edge length of the largest square for the
UraneSX, as well as the location of their respective centers. To
simplify the problem, the bounds on the velocity transmission
factors are such that $\psi_{min}=1/ \psi_{max}$.

\subsection{Regular dextrous workspace shape}
The Orthoglide and the UraneSX are compared according to the size
of their largest regular dextrous Workspace. Due to the
symmetrical architecture of the Orthoglide, the Cartesian
workspace has a fairly regular shape in which it is possible to
include a cube whose sides are parallel to the planes $xy$, $yz$
and $xz$ respectively. The Cartesian workspace of the UraneSX is
the intersection of three cylinders whose axes are parallel to the
$z$-axis. Thus, the workspace is unlimited in the $z$-direction
and the Jacobian matrix does not depend on the $z$ coordinate.
Only the limits on the linear joints define the limits of the
Cartesian workspace in the $z$-direction. However, it is possible
to include a square in the plane $xy$. Regular dextrous workspaces
are thus chosen to be a cube for the Orthoglide and a square for
the UraneSX.

\subsection{Geometric constraints}
\label{geometric_constraints} Section \ref{preliminaries} is
suggested to add geometrical constraints so as to decrease the
computing time and to avoid numerical problems on singularities.
Here, polynomial ${\cal P}$ is  defined only for the points within
the intersection ${\cal I}$ of the three cylinders defined by
 \begin{equation}
 x^2+y^2 < L^2 \quad
 x^2+z^2 < L^2 \quad
 y^2+z^2 < L^2
 \end{equation}
for the Orthoglide, and,
\begin{eqnarray}
 (x-R+r)^2+y^2 &<& L^2 \nonumber \\
 \left(x-(R-r)\frac{1}{2}\right)^2+\left(y-(R-r)\frac{\sqrt{3}}{2}\right)^2 &<& L^2 \nonumber \\
 \left(x-(R-r)\frac{1}{2}\right)^2+\left(y+(R-r)\frac{\sqrt{3}}{2}\right)^2 &<& L^2 \nonumber
\end{eqnarray}
for the UraneSX. With these constraints, matrix \negr B is never
singular and thus can be always inverted. To solve numerically the
above equations and to compare the two mechanisms, the length of
the legs is normalized, {\it i.e.} we set $L=1$.

\subsection{Comparison results}
\label{results}

To compare the two mechanisms studied, the leg length $L$ is set
to 1 and the bounds on the velocity factor amplification are set
to $\psi=[0.5~2]$, with $\alpha=0.001$. For the UraneSX, it is
necessary to define two additional lengths, $r$ and $R$. However,
the edge length of the workspace  depends only on $R-r$.

For the Orthoglide, it is found that the largest cube has its
center located at $(0.086, 0.086, 0.086)$, and that the cube edge
length is $L_{Workspace}= 0.644$. Also, using the incremental
approach described in section \ref{algorithm}, we get for the
Orthoglide the computation time of Table~\ref{tcw} on a Sun Blade
workstation.
\begin{table}[!hc]
  \begin{center}
\begin{tabular}{|c|c|c|c|c|}\hline
 Accuracy $\alpha$ (mm)& 0.01& 0.001& 0.0001& 0.00001\\ \hline
 Computation time (s) & 360 & 150 & 504 & 900 \\ \hline
\end{tabular}
\end{center}
\caption{\label{tcw} \bf Computation time of the largest cube
enclosed in the dextrous workspace for various accuracies}
\end{table}

For the UraneSX, the design parameters are those defined in
\cite{Company:2000}, which we have normalized to have $L=1$, {\it
i.e.} $r=3/26$ and $R= 7/13$. To compare the two mechanisms, we
increase the value of $R$ such that $R'=R+\lambda$ with
$\lambda=[0.0,0.2]$. For $R< 7/13$, the constraints on the
velocity amplification factors  are not satisfied.

\begin{table} [!hb]
  \begin{center}
   \begin{tabular}{|c|c|c|} \hline
$\lambda$ & Center & $L_{Workspace}$ \\ \hline
 0.00 &   (-0.0178,-0.0045)&   0.510\\ \hline
 0.05 &   (-0.0179,-0.0022)&   0.470\\ \hline
 0.10 &   (-0.0225,-0.0031)&   0.420\\ \hline
 0.15 &   (-0.0245,-0.0018)&   0.370\\ \hline
 0.20 &   (-0.0211,-0.0033)&   0.320\\ \hline
   \end{tabular}
   \caption{{\bf Variations of the edge length of the square workspace for the UraneSX mechanism}}
   \label{table_variation}
  \end{center}
\end {table}

The optimal value of $R'$ is obtained for $\lambda=0$, {\it i.e.}
for the design parameters defined in \cite{Company:2000} for an
industrial application (see table \ref{table_variation}). To
expand this square workspace in the $z$-direction, the range
limits must be equal to the edge length of the square plus the
range variations necessary to move throughout the square in the
$x-y$ plane.

The constraints on the velocity amplification factors used for the
design of the Orthoglide are close to those used for the design of
the UraneSX which is an industrial machine tool. For the same
length of the legs, the size of the cubic workspace is larger for
the Orthoglide than for the UraneSX.

For the Orthoglide, the optimization puts the serial and parallel
singularities far away from the Cartesian workspace
\cite{Chablat:2002}. The UraneSX has no parallel singularities due
to the design parameters ($R-r < L$), but serial singularities
cannot be avoided with the previous optimization function. To
produce the motion in the $z$-direction, the range limits of the
linear joints are set such that the constraints on the velocity
amplification factors are not satisfied throughout the Cartesian
workspace.

The range limits $\Delta \rho_i$ of each prismatic joint can be
decomposed into two parts. For the Orthoglide (resp. for the
UraneSX), the first part $\Delta f_i$ makes it possible to move
the mobile platform throughout the face of the prescribed cube
that is perpendicular to the considered prismatic joint axis
(resp. throughout the prescribed square). The second part is equal
to the edge length of the cubic workspace $L_{Workspace}$. The
equations of the inverse kinematic model allow us to compute
$\Delta f_i$ for the two mechanisms.

For the Orthoglide, the position and the size of the prescribed
cube define three range limits for the $x-y-z$ platform
coordinates,
 \bseq
  \beqa
     x&=& [-0.322+0.085, 0.322+0.085] \\
     y&=& [-0.322+0.085, 0.322+0.085] \\
     z&=& [-0.322+0.085, 0.322+0.085],
  \eeqa
For the UraneSX, the position and the size of the prescribed
square define two range limits for the $x-y$ platform coordinates,
 \beqa
       &x&= [-0.255-0.018,0.255-0.018] \\
       &y&= [-0.255,0.255].
 \eeqa
 \eseq
For the Orthoglide, all $\Delta f_i$ are equal due to the
symmetrical architecture. For the UraneSX, we take $\Delta f=
Max(\Delta f_i)$. The results are $\Delta f= 0.181$ and $\Delta
\rho=0.825$ for the Orthoglide and $\Delta f= 0.353$ and $\Delta
\rho=0.863$ for the UraneSX. This means that the range limits are
quite similar for the same leg length. To calculate the volume of
the Cartesian workspace of the two mechanisms for the previous
range limits, we have used a CAD system. Results are given in
table~\ref{table_dimensions}.
\begin{table}[hbt]
  \begin{center}
    \begin{tabular}{|c|c|c|c|} \hline
     ~~~~~~~~~~ & Cartesian workspace volume     & cubic dextrous workspace volume & ratio \\
     ~~~~~~~~~~ & with optimized ranges limits   &                                 &       \\ \hline
    Orthoglide  &          0.566  & 0.265        & 46.8\% \\ \hline
    UraneSX     &          0.544  & 0.132        & 24.3\% \\ \hline
   \end{tabular}
   \caption{{\bf Workspace volumes of the two mechanisms}}
   \protect\label{table_dimensions}
  \end{center}
\end {table}

\begin{figure}[hbt]
    \begin{center}
           \centerline{\hbox{\includegraphics[width=77mm,height=58mm]{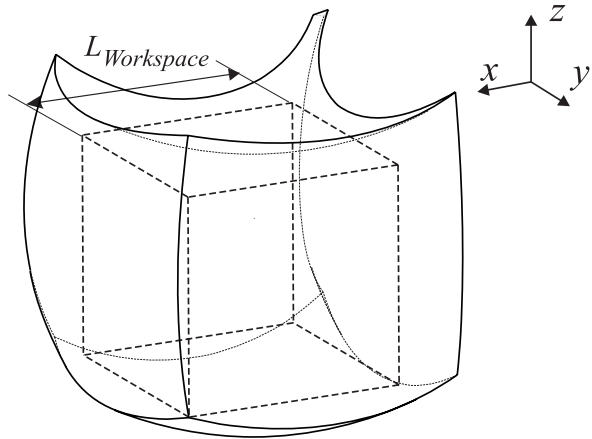}}}
           \caption{{\bf Cartesian workspace and dextrous workspace for the Orthoglide mechanism with optimized range limits}}
           \protect\label{figure:Orthoglide_Workspace}
    \end{center}
\end{figure}

\begin{figure}[hbt]
    \begin{center}
           \centerline{\hbox{\includegraphics[width=88mm,height=61mm]{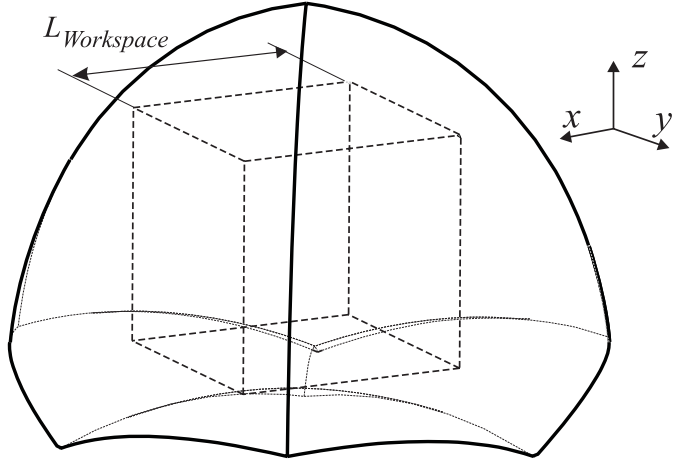}}}
           \caption{{\bf Cartesian workspace and dextrous workspace for the UraneSX mechanism with optimized range limits}}
           \protect\label{figure:UraneSX_Workspace}
    \end{center}
\end{figure}

To help  understand these results,
Fig.~\ref{figure:Orthoglide_Workspace} and
\ref{figure:UraneSX_Workspace} show the location of the largest
cubic workspace inside the Cartesian workspace. As the Cartesian
workspace of the Orthoglide is regular and admits a quasi-cubic
shape, the ratio between the cubic workspace and the Cartesian
workspace is better than for the UraneSX.

In table~\ref{table:synthesis}, the design parameters are compared
to achieve the same cubic dextrous workspace with
$L_{Workspace}=~1$. The legs length is directly connected to the
dynamic properties of the mechanism. The range limits and the legs
length are important parameters in the determination of the total
size of the mechanism and in its global cost. The volume of the
Cartesian workspace allows us to characterize the shape and the
volume of motion of the tool with regard to the useful Cartesian
workspace dedicated to manufacturing tasks (cubic workspace).
\begin{table}[hbt]
  \begin{center}
   \begin{tabular}{|c|c|c|c|} \hline
                 & Leg length & Range limits & Volume of the
                  Cartesian workspace                               \\ \hline
      Orthoglide & 1.55       & 1.28         & 2.13                 \\ \hline
      UraneSX    & 1.96       & 1.69         & 4.12                 \\ \hline
   \end{tabular}
   \caption{{\bf Synthesis of the comparative study for the same cubic Cartesian workspace}}
   \label{table:synthesis}
  \end{center}
\end {table}

These criteria allow us to optimize some geometric parameters to
design a machine tool for milling applications. Although in this
approach, the kinetostatic properties of the Orthoglide are better
than the UraneSX ones, we cannot assert that the Orthoglide is
better than the UraneSX. One reason is that these two PKM are not
aimed at identical manufacturing tasks. The main applications of
the UraneSX are drilling, facing and tapping whereas the
Orthoglide is more universal.

Other shapes of regular dextrous workspaces can be computed for
the Orthoglide and the UraneSX by using cylindrical or spherical
coordinates to have the largest cylinder or sphere respectively,
even if these shapes are generally less relevant for milling
applications.

\section{Conclusions}
This paper introduces an interval analysis based study for the
design and the comparison of 3-DOF PKM. Two basic tools and an
algorithm are described to determine the largest regular dextrous
workspace enclosed in the Cartesian workspace. The dextrous
workspace is a part of the Cartesian workspace in which the
velocity amplification factors remain within a predefined range.
This means that throughout the dextrous workspace, milling tool
paths are available because the variations of the kinematic
performances index remain under reasonable values. The regular
dextrous workspace shape is a cube for the Orthoglide and a square
for the UraneSX. This general method is coupled with geometric
constraints associated with the mechanisms studied to avoid
numerical problems at singular configurations. The shape of the
dextrous workspace was chosen for milling applications but it can
be different for other applications. The range limits and the
volume of the Cartesian workspace were calculated to compare the
two mechanisms.
\section{Acknowledgments}
This research was supported by the CNRS (Project ROBEA ``Machine à
Architecture compleXe''). We would like to thank the anonymous
reviewers for their very useful comments.
\def\refname{\large References}
\bibliographystyle{unsrt}

\end{document}